\documentclass[nohyperref]{article}

\usepackage{microtype}
\usepackage{graphicx}
\usepackage{subfigure}
\usepackage{booktabs} % for professional tables
\usepackage{multirow}

\usepackage{hyperref}

\usepackage[accepted]{icml2022}

\usepackage{amsmath}
\usepackage{amssymb}
\usepackage{mathtools}
\usepackage{amsthm}

\usepackage[capitalize,noabbrev]{cleveref}

\theoremstyle{plain}

\theoremstyle{definition}

\theoremstyle{remark}

\usepackage[textsize=tiny]{todonotes}

\icmltitlerunning{}

\begin{document}

\twocolumn[
\icmltitle{MGC: A Complex-Valued Graph Convolutional Network for Directed Graphs}



\icmlsetsymbol{equal}{*}

\begin{icmlauthorlist}
\icmlauthor{Jie Zhang}{ncu}
\icmlauthor{Bo Hui}{au}
\icmlauthor{Po-Wei Harn}{au}
\icmlauthor{Min-Te Sun}{ncu}
\icmlauthor{Wei-Shinn Ku}{au}
\end{icmlauthorlist}

\icmlaffiliation{ncu}{Department of Computer Science and Information Engineering, National Central University, Taiwan.}
\icmlaffiliation{au}{Department of Computer Science and Software Engineering, Auburn University, United States}

\icmlkeywords{Machine Learning, ICML}

\vskip 0.3in
]



\printAffiliationsAndNotice{}  % leave blank if no need to mention equal contribution

\begin{abstract}
Recent advancements in Graph Neural Networks have led to state-of-the-art performance on graph representation learning. However, the majority of existing works process directed graphs by symmetrization, which causes loss of directional information. To address this issue, we introduce the magnetic Laplacian, a discrete Schr{\"o}dinger operator with magnetic field, which preserves edge directionality by encoding it into a complex phase with an electric charge parameter. By adopting a truncated variant of PageRank named LinearRank, we design and build a low-pass filter for homogeneous graphs and a high-pass filter for heterogeneous graphs. In this work, we propose a complex-valued graph convolutional network named \underline{M}agnetic \underline{G}raph \underline{C}onvolutional network (MGC). With the corresponding complex-valued techniques, we ensure our model will be degenerated into real-valued when the charge parameter is in specific values. We test our model on several graph datasets including directed homogeneous and heterogeneous graphs. The experimental results demonstrate that MGC is fast, powerful, and widely applicable.
\end{abstract}

\section{Introduction}
High dimensional data are ordinarily represented as graphs in a variety of areas, such as citation, energy, sensor, social, and traffic networks. Consequently, Graph Neural Networks (GNNs), also known as CNNs on graphs, have shown tremendous promise. 

Since the emergence of GCN~\cite{DBLP:conf/iclr/KipfW17}, numerous GNNs have been developed to generalize convolution operations on graphs through the graph Fourier transform. Graph convolution is based on graph signal processing~\cite{6494675,SHUMAN2016260}, where the graph filter is a crucial component. A graph filter is a tensor which processes a graph signal by amplifying or attenuating its corresponding graph Fourier coefficients. Generally, a graph filter is a combination of different powers of a normalized adjacency matrix or a normalized Laplacian matrix. 

The graph Fourier transform, which is based on computing the eigendecomposition of a graph filter, cannot be directly applied to a directed graph, since the adjacency matrix is asymmetric~\cite{TREMBLAY2018299,Stankovic2019}. A straightforward method is to symmetrize the adjacency matrix which treats a directed graph as an undirected graph. This operation is adopted by most of the existing GNNs. Unfortunately, it overlooks the asymmetric nature which leads to the situation that GNNs fail to capture the directional information on directed graphs. An alternative solution is Chung's directed Laplacian~\cite{chung2005laplacians} which is based on the transition probability matrix of out-degree and the corresponding Perron vector. It is one of the directed Laplacians which takes advantage of the directional information hidden behind the transition probability matrix of out-degree.

\begin{figure}[t]
    \centering
    \vspace{-4mm}
    \includegraphics[width=0.9\linewidth]{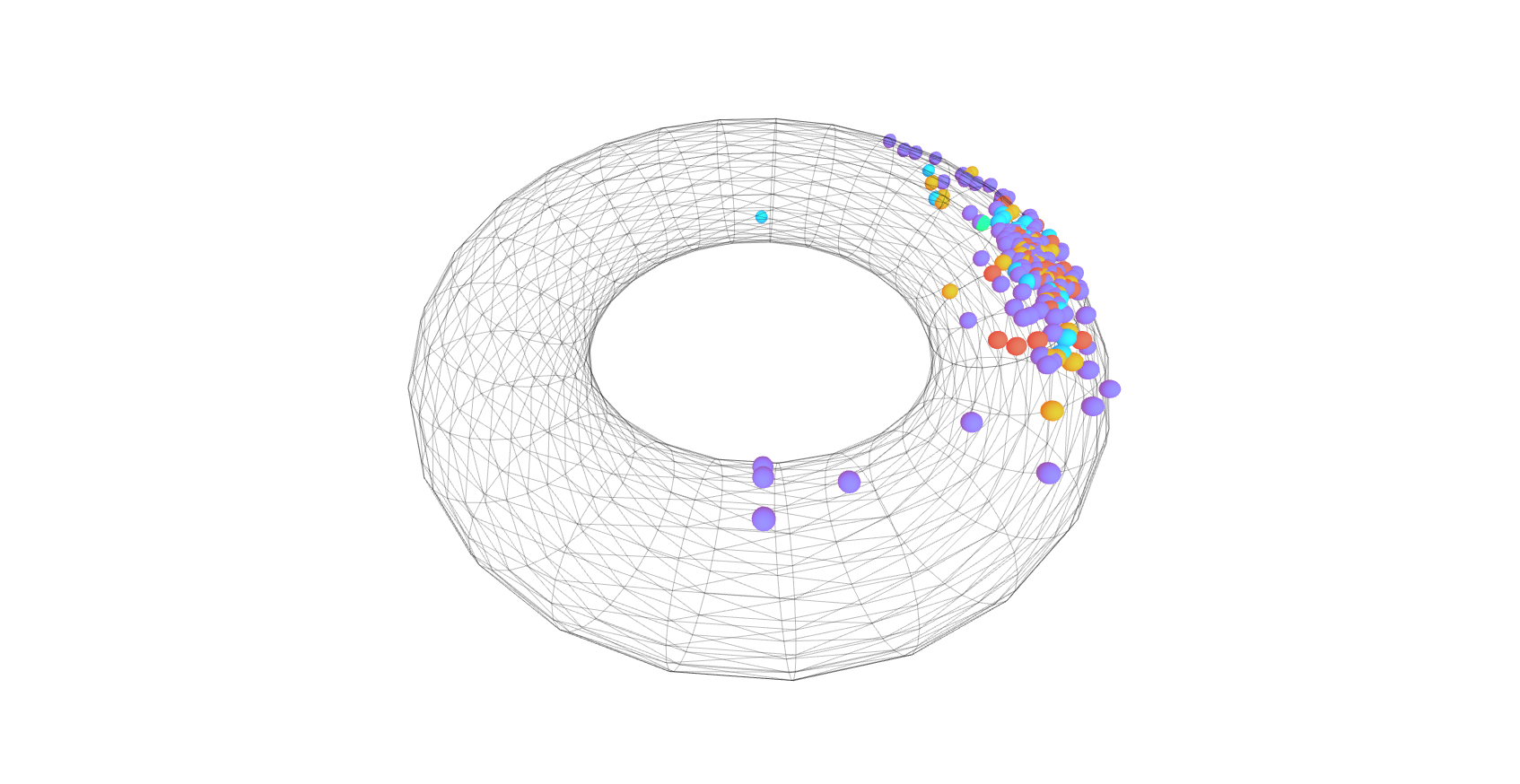}
    \vspace{-8mm}
    \caption{The visualization of the magnetic eigenmaps with the electric parameter $q = \frac{1}{3}$ in Texas. Each node represents an eigenvector.}
    \label{fig:magnetic_eigenmaps}
    \vspace{-6mm}
\end{figure}

In this paper, we introduce another directed Laplacian named the magnetic Laplacian~\cite{PhysRevE.95.022302,FANUEL2018189} which utilizes the directional information from the adjacency matrix. As an extension of the combinatorial Laplacian~\cite{chung1997spectral}, the magnetic Laplacian preserves edge directionality by encoding it into a complex phase. We adopt the magnetic Laplacian for node classification in directed graphs. Figure~\ref{fig:magnetic_eigenmaps} exhibits the visualization of the magnetic eigenmaps~\cite{FANUEL2018189}, which are eigenvectors of the magnetic Laplacian, with the electric parameter $q = \frac{1}{3}$ in Texas~\cite{lu2003link}.

Recently, enlightened by spectral ranking~\cite{vigna_2016}, GNNs such as PPNP, APPNP~\cite{klicpera2018combining}, GDC~\cite{NEURIPS2019_23c89427}, and S$^{2}$GC~\cite{zhu2021simple} demonstrate strong performance in node classification. In this research, we introduce a truncated variant of PageRank named LinearRank~\cite{10.1145/1148170.1148225}. By combining LinearRank with the renormalized adjacency matrix, we design a low-pass filter for homogeneous graphs. Furthermore, by associating LinearRank with the negative renormalized adjacency matrix, we design a high-pass filter for heterogeneous graphs. In short, we propose a complex-valued graph convolutional network named Magnetic Graph Convolutional network (MGC). The experimental results demonstrate that MGC is fast, powerful, and widely applicable.
\section{Related Work}
\noindent\textbf{Graph Filters and Spectral Ranking.}
Based on graph signal processing, \cite{DBLP:conf/nips/DefferrardBV16} proposes a fast localized graph filter. ChebyNet~\cite{DBLP:conf/nips/DefferrardBV16} is the first GNN with graph filter modification, which utilizes the Chebyshev polynomials of the first kind. Then GCN~\cite{DBLP:conf/iclr/KipfW17} improves ChebyNet by proposing a method called the renormalization trick which inspires several GNNs to design filters. As a simplified version of multi-layer GCN, SGC~\cite{pmlr-v97-wu19e} eliminates nonlinear activation functions and Dropout in order to retain performance and achieve the same results as GCN.

Since graph convolution is related to graph signal processing, several researchers have tried to design GNNs from the view of devising graph filters. One common approach is focusing on spectral ranking~\cite{vigna_2016}. Personalized PageRank~\cite{10.1145/775152.775191} is a well-known spectral ranking algorithm adopted by PPNP and APPNP~\cite{klicpera2018combining}. Subsequently, GDC~\cite{NEURIPS2019_23c89427} utilizes Personalized PageRank and Heat Kernel PageRank~\cite{Chung19735} to construct a general graph convolution framework. Afterwards, S$^{2}$GC~\cite{zhu2021simple} adopts Markov Diffusion kernel~\cite{4053117} as its graph filter.

\noindent\textbf{Chung's Directed Laplacian.}
Chung's directed Laplacian~\cite{chung2005laplacians} is a Hermitian matrix based on the transition probability matrix of out-degree and the corresponding Perron vector. This Laplacian is applied for strongly connected graphs. \cite{ma2019spectralbased} adopts this directed Laplacian and combines it with the renormalization trick to propose DGCN for directed graphs. The work in \cite{NEURIPS2020_cffb6e22} inherits DGCN, and generalizes it for all kinds of graphs with PageRank~\cite{ilprints422} and Personalized PageRank.

\noindent\textbf{Magnetic Laplacian.}
The magnetic Laplacian is first proposed in \cite{shubin1994discrete}. As a discrete Hamiltonian of a charged particle on a graph, the magnetic Laplacian is widely used in mathematics~\cite{colin2013magnetic} and physics~\cite{Olgiati2017}. \cite{biamonte2019complex} indicates that the magnetic Laplacian is deeply related with Quantum walk. Since it is a complex Hermitian matrix, its eigenvalues are real and eigenvectors are orthonormal. Due to these properties, the magnetic Laplacian can be employed in graph representation learning for both directed and undirected graphs.

We are aware of a concurrently developed work: MagNet~\cite{zhang2021magnet}, which adopts the renormalized magnetic adjacency matrix as the graph shift operator~\cite{Stankovic2019} to build a graph filter. It is composed of two layers of graph convolution for node classification. MagNet separates complex-valued tensors into the real and imaginary parts, each of which is processed by an independent real-valued neural network (RVNN). In the last layer, there is a concatenation operation to turn a complex-valued tensor into real-valued. To determine the electric parameter $q$ of the magnetic Laplacian, binary search is adopted.

Our model is different from MagNet in several ways. First, we utilize Johnson's algorithm~\cite{doi:10.1137/0204007} for determining $q$. Second, we adopt LinearRank with the renormalized magnetic adjacency matrix for homogeneous graphs and the negative renormalized magnetic adjacency matrix for heterogeneous graphs. Third, we adopt a complex-valued graph convolution with corresponding weight initialization~\cite{trabelsi2018deep}. Fourth, inspired by gating mechanism~\cite{pmlr-v70-dauphin17a}, we design a complex-valued activation function to turn a complex-valued tensor into real-valued. Notice when $q = 0$ or $q = \frac{1}{2}$, our model will be degenerated from a complex-valued neural network (CVNN) to an RVNN.
\section{Preliminary}
\subsection{Directed Graphs and Undirected Graphs}
A directed graph $G$ is represented as $G = \{\mathcal{V},\mathcal{E}\}$, where $\mathcal{V} = \{v_{0},v_{1},...,v_{n-1}\}$ is a finite set of vertices with $|\mathcal{V}| = n$, and $\mathcal{E} \subseteq \mathcal{V}\times\mathcal{V}$ is the set of edges. Let $\mathbf{A} \in \mathbb{R}^{n \times n}$ denote the directed adjacency matrix of $G$, where $\mathbf{A}(u,v) = 1$ if there is an edge from node ${u}$ to node ${v}$, otherwise $\mathbf{A}(u,v) = 0$.

In graph theory, directed graphs can be divided into two categories: directed acyclic graphs and directed cyclic graphs. We define a cycle in a directed graph as a closed chain of distinct edges that connects a sequence of distinct nodes. If all the edges on a cycle are oriented in the same direction, it is called a directed cycle. If a directed graph has no directed cycle, we name it a directed acyclic graph. Otherwise, the directed graph is called directed cyclic graph.

We use $G_{s}=\{\mathcal{V},\mathcal{E}_{s}\}$ to represent an undirected graph. The undirected adjacency matrix can be described by $\mathbf{A}_{s} = \frac{1}{2}(\mathbf{A} + \mathbf{A}^{\mathrm{T}})$ or $\mathbf{A}_{s}(u,v) = \mathrm{max}(\mathbf{A}(u,v), \mathbf{A}(v,u))$, so that $\mathbf{A}_{s}$ is symmetric. For an undirected graph, the combinatorial Laplacian~\cite{chung1997spectral} can be defined as $\mathbf{L} = \mathbf{D}_{s} - \mathbf{A}_{s}$, where $\mathbf{D}_{s} \in \mathbb{R}^{n \times n}$ is the diagonal degree matrix of $G_{s}$.

\subsection{Homogeneous Graphs and Heterogeneous Graphs}
Graphs can be either homogeneous or heterogeneous. The homophily and heterophily of a graph are used to describe the relation of labels among nodes. A homogeneous graph is a graph where the labels of all the nodes are consistent. On the contrary, in a heterogeneous graph, labels for nodes are of different types. The node homophily index for a graph~\cite{Pei2020Geom-GCN} is denoted as $\mathcal{H}_{\mathrm{node}}(G) = \frac{1}{|\mathcal{V}|}\sum_{u \in \mathcal{V}}{\frac{\left|\{v|v\in\mathcal{N}_{u},\mathcal{Y}_{v}=\mathcal{Y}_{u}\}\right|}{|\mathcal{N}_{u}|}}$, where $\mathcal{N}_{u}$ is the neighbor set of node $u$ and $\mathcal{Y}_{u}$ is the label of node $u$. Note that $\mathcal{H}_{\mathrm{node}}(G) \rightarrow 1$ indicates strong homophily and $\mathcal{H}_{\mathrm{node}}(G) \rightarrow 0$ corresponds to strong heterophily. 

\subsection{Graph Signal}
In graph signal processing (GSP), a graph signal~\cite{Stankovic2019} $x$ is a map from the set of vertices $V$ into the set of complex numbers $\mathbb{C}$ as
$x:\ \mathcal{V} \rightarrow \mathbb{C};\ v_{n} \rightarrow x_{n}$. For convenient mathematical representation, we denote graph signals as a column vector $\mathbf{x} = [x_{0},x_{1},...,x_{n-1}]^{\mathrm{T}}$.

\subsection{Graph Shift and Graph Shift Operator}
A graph shift operator (GSO)~\cite{Stankovic2019} is a matrix $\mathbf{S} \in \mathbb{C}^{n \times n}$ where $\mathbf{S}{(u, v)} \neq {0}$ if and only if $u = v$ or $(u,v) \in \mathcal{E}$. A GSO defines how to shift a graph signal from one node to its neighbors based on the graph topology. More specifically, GSO is a local operator that replaces graph signal value of each node with linear combination of its one-hop neighbors'. In graph signal processing, it is common to take a normalized adjacency matrix or normalized Laplacian matrix as a graph shift operator.

\subsection{Graph Filter and Graph Frequency Response Function}
A graph filter $\mathbf{H} \in \mathbb{C}^{n \times n}$~\cite{Stankovic2019} is a function $h(\cdot)$ of a graph shift operator, denoted as $\mathbf{H} = h(\mathbf{S})$. Apparently, a graph shift operator is a simple graph filter. In graph signal processing, it is common to utilize a polynomial graph filter which is defined as $h(\mathbf{S}) = \sum^{K}_{k=0}{{\zeta}_{k}\mathbf{S}^{k}}$. This kind of graph filter is named Moving-Average (MA) filter~\cite{7581108}, which is also named Finite Impulse Response (FIR) filter. The capacity of a GNN with an FIR filter is determined by the degree $k$ of a polynomial. We denote it as a MA$_{K}$ or FIR$_{K}$ filter. Compared with a FIR filter, an Infinite Impulse Response (IIR) filter is more specialized to capture global structure on a graph. A well-known IIR filter is called Auto-Regressive (AR) filter~\cite{7581108}, which is defined as $h(\mathbf{S}) = \left(\mathbf{I}_{n}+\sum^{K}_{k=1}{{\eta}_{k}\mathbf{S}^{k}}\right)^{-1}$. We denote it as AR$_{K}$ filter. 

By multiplying an AR$_{K}$ filter with an MA$_{K}$ filter, we can obtain an Auto-Regressive–Moving-Average (ARMA) filter~\cite{7581108} of order $K$, denoted as ARMA$_{K}$. It is defined as $h(\mathbf{S}) = \left(\sum^{K}_{k=0}{{\zeta}_{k}\mathbf{S}^{k}}\right)\left(\mathbf{I}_{n} + \sum^{K}_{k=1}{{\eta}_{k}\mathbf{S}^{k}}\right)^{-1}$. With the same order $K$, an ARMA$_{K}$ filter has better performance than an AR$_{K}$ filter or an MA$_{K}$ filter, since an ARMA$_{K}$ filter is composed of a feedforward term (MA) and a feedback term (AR). An ARMA$_{K}$ filter is a special case of an ARMA$_{F,B}$ filter as $h(\mathbf{S}) = \left(\sum^{F}_{f=0}{{\zeta}_{f}\mathbf{S}^{k}}\right)\left(\mathbf{I}_{n} + \sum^{B}_{b=1}{{\eta}_{b}\mathbf{S}^{k}}\right)^{-1}$.

A function $h(\cdot)$ of the set of the graph filter’s eigenvalues $\lambda$, $h(\lambda)$ is called the graph frequency response function (GFRF), which extends the convolution theorem from digital signal processing to graphs. For a normalized Laplacian matrix, its GFRF is $h(\lambda) = \lambda$. For a normalized adjacency matrix, its GFRF is $h(\lambda) = 1 - \lambda$. In graph signal processing, the reason to choose a normalized Laplacian matrix as graph filter is because its graph filter is the simplest high-pass filter. Likewise, a normalized adjacency matrix is the simplest low-pass filter.

\subsection{Graph Fourier Transform}
For an undirected graph $G_s$, its graph filter can be eigendecomposed as $\mathbf{H} = \bf{U}\bf{\Lambda}\mathbf{U}^{*}$, where $\mathbf{U} \in \mathbb{R}^{n \times n}$ is a matrix of orthonormal eigenvectors, $\mathbf{\Lambda} = \mathrm{diag}\left([h(\lambda^{(0)}),...,h(\lambda^{(n-1)})]\right) \in \mathbb{R}^{n \times n}$ is a diagonal matrix of filtered eigenvalues, and ${*}$ means conjugate transpose. Since $\mathbf{U}$ is real-valued, we have $\mathbf{U}^{*} = \mathbf{U}^{\mathrm{T}}$.

Based on the theory of graph signal processing, the graph Fourier transform for a signal vector $\mathbf{x}$ on an undirected graph $G_s$ is defined as $\hat{\mathbf{x}} = \mathbf{U}^{\mathrm{T}}\mathbf{x}$, and the inverse graph Fourier transform is $\mathbf{x} = \mathbf{U}\hat{\mathbf{x}}$. The convolution operator on graph $*_{\mathcal{G}}$ is defined as $\mathbf{h}*_{\mathcal{G}}\mathbf{x} = \mathbf{U}\left((\mathbf{U}^{{T}}\mathbf{h})\odot(\mathbf{U}^{{T}}\mathbf{x})\right)=\mathbf{U}{h{(\mathbf{\Lambda})}}\mathbf{U}^{\mathrm{T}}\mathbf{x}=\mathbf{H}\mathbf{x}=\sum_{i=0}^{n-1}{h(\lambda^{(i)})\mathbf{u}_{i}\mathbf{u}_{i}^{\mathrm{T}}}$, where $\mathbf{h}$ is a vector form of GFRF, ${h{(\mathbf{\Lambda})}}$ is a matrix form of GFRF, and $\mathbf{u}_{0},\mathbf{u}_{1},...,\mathbf{u}_{n-1}$ are eigenvectors of $\mathbf{U}$.
\section{Proposed Method}
\subsection{Magnetic Laplacian and Directed Cycles}
For a directed graph, its adjacency matrix is asymmetric. Consequently, the combinatorial Laplacian is not straightforwardly applicable. To process a directed graph while preserving edge directionality, we introduce the magnetic Laplacian~\cite{PhysRevE.95.022302,FANUEL2018189}, which is a complex positive semi-definite Hermitian matrix. As a deformation of the combinatorial Laplacian, the magnetic Laplacian is defined as:
\begin{equation}
\centering
\begin{split}
\mathbf{L}_{q} &= \mathbf{D}_{s} - \mathbf{A}_{s}\odot{\mathbf{T}_{q}},\\
\mathbf{T}_{q}{(u,v)} &= {\exp}{\left(i2{\pi}q{\left(\mathbf{A}(u,v) - \mathbf{A}(v,u)\right)}\right)},
\end{split}
\end{equation}
where $\odot$ represents the Hadamard product, $\mathbf{T}_{q}$ is a complex and unitary parallel transporter which represents the direction of magnetic fluxes among nodes, and $q$ is an electric charge parameter. The symmetric normalized magnetic Laplacian is defined as
\begin{equation}
\centering
\begin{split}
\mathcal{L}_{q} &= \mathbf{D}^{-\frac{1}{2}}_{s}\left(\mathbf{D}_{s} - \mathbf{A}_{s}\odot{\mathbf{T}_{q}}\right)\mathbf{D}^{-\frac{1}{2}}_{s}\\
&= \mathbf{I}_{n} - \mathbf{D}^{-\frac{1}{2}}_{s}\mathbf{A}_{s}\mathbf{D}^{-\frac{1}{2}}_{s}\odot{\mathbf{T}_{q}}.
\end{split}
\end{equation}
Since $q$ has a high influence for the magnetic Laplacian, we should prudentially choose a specific value. Unfortunately, it is a tough issue because we only know that $q$ is closely related to reciprocity and directed cycles of a graph. To address this issue, \cite{PhysRevE.95.022302} proposes a method which is $q = \frac{1}{m}$, if there is a directed $m$-cycle (where $m \ge 2$) in a graph. Note that a reciprocal link is treated as a directed 2-cycle. $q = 0$ for both directed acyclic graphs and undirected graphs. Besides, to avoid the emphasis of reciprocal links, $q$ should be restricted in $[0, \frac{1}{2}]$. 

We introduce Johnson's algorithm~\cite{doi:10.1137/0204007} to determine directed cycles for a directed cyclic graph. The time complexity is $O\left(|\mathcal{V}|^{2}\cdot\log{|\mathcal{V}|} + |\mathcal{V}|\cdot|\mathcal{E}|\right)$. Once we have detected all cycles, we will successively choose $q$ from a set of reciprocals of cycles' length. Notice that there are three special cases when the magnetic Laplacian will be degenerated due to the value of the electric charge parameter: (1)~$q = 0$, i.e., the combinatorial Laplacian. It means that the graph is an undirected graph or a directed acyclic graph. (2)~$q = \frac{1}{4}$, i.e., the imaginary combinatorial Laplacian. (3)~$q = \frac{1}{2}$, i.e., the signed Laplacian~\cite{doi:10.1137/1.9781611972801.49}. It means that the graph is composed of both positive and negative edges.

\subsection{Magnetic Laplacian v.s. Chung's Directed Laplacian}
Chung's directed Laplacian~\cite{chung2005laplacians} is defined as $\mathbf{L}_{+} = \mathbf{\Phi} - \frac{1}{2}\left(\mathbf{\Phi}\mathbf{P}_{\mathrm{out}} + \mathbf{P}^{*}_{\mathrm{out}}\mathbf{\Phi}\right)$, where $\mathbf{P}_{\mathrm{out}}$ is the transition probability matrix of out-degree, $\mathbf{\Phi}$ is the diagonal matrix with entries $\mathbf{\Phi}(v, v) = \mathbf{\phi}(v)$, and $\mathbf{\phi}$ is the Perron vector of $\mathbf{P}_{\mathrm{out}}$. For an undirected graph, $\mathbf{\Phi} = \mathbf{D}$, thus $\mathbf{L}_{+}$ will be 
degenerated to $\mathbf{L}$. The normalized Chung's directed Laplacian is defined as $\mathcal{L}_{+} = \mathbf{I}_{n} - \frac{1}{2}\left(\mathbf{\Phi}^{\frac{1}{2}}\mathbf{P}_{\mathrm{out}}\mathbf{\Phi}^{-\frac{1}{2}} + \mathbf{\Phi}^{-\frac{1}{2}}\mathbf{P}^{*}_{\mathrm{out}}\mathbf{\Phi}^{\frac{1}{2}}\right)$.

Compared with Chung's directed Laplacian, the magnetic Laplacian has several superiorities. First, the normalized magnetic adjacency matrix is likely to be a sparse matrix unless the graph itself is fully connected. It is because that the normalized magnetic adjacency matrix is the element-wise production of the normalized adjacency matrix and the parallel transpose matrix, where the normalized adjacency matrix is calculated from the unnormalized adjacency matrix and the degree matrix. The normalized adjacency matrix is likely to be sparse because the degree matrix is always sparse and the unnormalized adjacency matrix is also sparse unless the graph is fully connected. As a consequence, the normalized magnetic adjacency matrix is likely to be a sparse matrix no matter if the parallel transpose matrix is sparse or not.

Second, to apply Chung's directed Laplacian, there is an assumption that the graph is a strongly connected graph, which means a graph cannot contain isolated nodes or be a bipartite graph. Because the calculation process of Chung's directed Laplacian requires a Perron vector of the transition probability matrix, in which every element has to be positive. Otherwise, the transition probability matrix will not be irreducible and aperiodic, and it may not have a unique stationary distribution. Although in \cite{NEURIPS2020_cffb6e22} two approaches are proposed to ensure the calculation of Chung's directed Laplacian. Inevitably, the normalized Chung's directed adjacency matrix is dense due to their calculation method.

\subsection{Graph Diffusion Filter}
Generally, a graph diffusion filter is defined as 
\begin{equation}
\centering
\begin{split}
\mathbf{H} = {\sum}^{\infty}_{k=0}{\theta}_{k}\mathbf{P}^{k},
\end{split}
\end{equation}
where ${\theta}_{k}$ is a damping factor with ${\sum}^{\infty}_{k=0}{\theta}_{k}=1$, and $\mathbf{P}$ is the transition probability matrix as a GSO. This diffusion filter belongs to a spectral ranking algorithm called Generalized PageRank (GPR)~\cite{10.1145/1148170.1148225,doi:10.1137/140976649,NEURIPS2019_9ac1382f}. Defining damping factors by a function named damping function as $\theta_{k}=f(k)$, we could utilize a variety of series to obtain convergent diffusion filters.

\subsubsection{Personalized PageRank, Heat Kernel PageRank, and Markov Diffusion}
A well-known example is Personalized PageRank (PPR)~\cite{10.1145/775152.775191}. Given a damping function $f(k) = {(1-\alpha)}{\alpha}^{k}$, then the diffusion filter of PPR is defined as
\begin{equation}
\centering
\begin{split}
\mathbf{H}_{\mathrm{PPR}} &= {\sum}^{\infty}_{k=0}{(1-\alpha)}{\alpha}^{k}\mathbf{P}^{k}\\
&={(1-\alpha)}\left(\mathbf{I}_{n} - {\alpha}\mathbf{P}\right)^{-1},\\  
\end{split}
\end{equation}
where ${\alpha} \in (0,1)$ is the restart probability, $\mathbf{I}_{n} - {\alpha}\mathbf{P}$ is not singular, and $\mathbf{P} \neq \frac{1}{\alpha}\mathbf{I}_{n}$. It is an ARMA$_{0,1}$ filter.

Another common instance is Heat Kernel PageRank (HKPR)~\cite{Chung19735}. Given a damping function $f(k) = \frac{\exp{(-t)}t^k}{k!}$, then we obtain the filter as
\begin{equation}
\centering
\begin{split}
\mathbf{H}_{\mathrm{HKPR}} &= {\sum}^{\infty}_{k=0}\frac{\exp{(-t)}{t}^{k}}{k!}\mathbf{P}^{k}\\
&= \exp{\left(-t(\mathbf{I}_{n} - \mathbf{P})\right)},\\
\end{split}
\end{equation}
where $t \in (0, \infty)$ is the diffusion time. Because the range of $t$ is an open interval, if $t$ is not controlled in a reasonable closed interval, it is hard to apply HKPR to realistic datasets.

In practice, infinite series based variants have to be truncated. It is general to calculate a filter of infinite series based variant by approximants, such as Taylor series, Chebyshev polynomial~\cite{HAMMOND2011129}, Bernstein polynomial~\cite{he2021bernnet}, and Padé approximant~\cite{PEROTTI201895}, as the form of an MA$_{K}$ filter, like $\sum^{K}_{k=0}\gamma_{k}\mathbf{P}^{k}$ with
\begin{equation}
\centering
\begin{split}
\left\lVert{{\sum}^{\infty}_{k=0}{\theta}_{k}\mathbf{P}^{k} - \sum^{K}_{k=0}\gamma_{k}\mathbf{P}^{k}}\right\rVert_{p} \le \epsilon,
\end{split}
\end{equation}
where $\epsilon$ is a round-off error. Instead of adopting a diffusion filter with an infinite order, we would like to focus on a kernel of truncated variants of GPR due to approximation-free. A simple truncated diffusion filter is named Markov Diffusion (MD) kernel~\cite{4053117}, which is defined as
\begin{equation}
\centering
\begin{split}
\mathbf{H}_{\mathrm{MD}} &= \sum^{K}_{k=1}{\frac{1}{K}\mathbf{P}^{k}} \\
&= \frac{1}{K}\mathbf{P}\left(\mathbf{I}_{n} - \mathbf{P}^{K}\right)\left(\mathbf{I}_{n}-\mathbf{P}\right)^{-1},
\end{split}
\end{equation}
where $\mathbf{I}_{n}-\mathbf{P}$ is not singular, and $\mathbf{P} \neq \mathbf{I}_{n}$. It is an ARMA$_{K+1,1}$ filter.

\subsubsection{LinearRank}
Apparently, Markov Diffusion kernel is not appropriate due to its fixed damping function. There is no decay between each vertex's $k$-hop neighbors and $(k+1)$-hop neighbors. To improve the Markov Diffusion kernel, we introduce LinearRank~\cite{10.1145/1148170.1148225}. The filter of LinearRank (LR) is defined as
\begin{equation}
\centering
\begin{split}
\mathbf{H}_{\mathrm{LR}} &= \sum^{K-1}_{k=0}{\frac{2(K-k)}{K(K+1)}\mathbf{P}^{k}} \\
&=\frac{2}{K(K+1)}\left(K\mathbf{I}_{n}-(K+1)\mathbf{P}+\mathbf{P}^{K+1}\right)(\mathbf{I}_{n}-\mathbf{P})^{-2},
\end{split}
\end{equation}
where $\mathbf{I}_{n}-\mathbf{P}$ is not singular and $\mathbf{P} \neq \mathbf{I}_{n}$. It is an ARMA$_{K+1,2}$ filter. For homogeneous graphs, we adopt the symmetric renormalized adjacency matrix as the GSO: $\mathbf{P} = \widetilde{\mathcal{A}}_{q} = \widetilde{\mathbf{D}}^{-\frac{1}{2}}_{s}\widetilde{\mathbf{A}}_{s}\widetilde{\mathbf{D}}^{-\frac{1}{2}}_{s}\odot{\mathbf{T}_{q}}$, where $\widetilde{\mathbf{A}}_{s} = \mathbf{A}_{s} + \mathbf{I}_{n}$, $\widetilde{\mathbf{D}}_{s} = \sum_{v}\widetilde{\mathbf{A}}_{s}{(u, v)}$. For heterogeneous graphs, we adopt the negative symmetric renormalized magnetic adjacency matrix as the GSO: $\mathbf{P} = -\widetilde{\mathcal{A}}_{q}$.

\subsection{Graph Signal Denoising}
Given a noisy graph signal vector $\mathbf{x} = \mathbf{\bar{x}} + \mathbf{n}$ with a pure part $\mathbf{\bar{x}}$ and a noisy part $\mathbf{n}$, we formulate graph signal denoising as
\begin{equation}\label{eqn:gsd}
\centering
\begin{split}
\operatorname*{min}_{\mathbf{x}} \{\mu\left\lVert{\mathbf{\bar{x}}-\mathbf{x}}\right\rVert^{2}_{2} + \mathcal{S}_{p}(\mathbf{x})\},
\end{split}
\end{equation}
where $\mu > 0$ is a trade-off coefficient and $\mathcal{S}_{p}(\mathbf{x})$ is the Dirichlet energy~\cite{6494675,cai2020a} over an undirected graph as
\begin{equation}
\centering
\begin{split}
\mathcal{S}_{p}(\mathbf{x}) &= \frac{1}{p}\sum_{u \in \mathcal{V}}\left\lVert{\nabla_{u}\mathbf{x}}\right\rVert^{p}_{2} \\
&= \frac{1}{p}\sum_{u \in \mathcal{V}}\left[\sum_{v \in \mathcal{N}_{u}}{\mathbf{A}_{s}(u,v)}[x(u)-x(v)]^{2}\right]^{\frac{p}{2}} \\
&= \frac{1}{p}\sum_{(u,v) \in \mathcal{E}}{\mathbf{A}_{s}^{\frac{p}{2}}(u,v)}[x(u)-x(v)]^{p}.
\end{split}
\end{equation}
When $p = 1$, $\mathcal{S}_{1}(\mathbf{x}) = \sqrt[2]{\mathbf{A}_{s}(u,v)}[x(u)-x(v)] = [\nabla\mathbf{x}](u,v)$ is the graph gradient. When $p = 2$, $\mathcal{S}_{2}(\mathbf{x}) = \frac{1}{2}\sum_{(u,v) \in \mathcal{E}}\mathbf{A}_{s}(u,v)[x(u)-x(v)]^{2}=\mathbf{x}^{*}\mathcal{L}\mathbf{x}$ is the graph Laplacian quadratic form. Besides, the $p$-Laplacian~\cite{lindqvist2019notes} can be derived from $\mathcal{S}_{p}(\mathbf{x})$.

Since in graph signal processing, the normalized Laplacian is frequently adopted as a GSO and the normalized Laplacian is a special case of the normalized magnetic Laplacian, we choose to discuss the condition when $p = 2$. When $p = 2$, $\mathcal{S}_{2}(\mathbf{x}) = \mathbf{x}^{*}\mathcal{L}\mathbf{x}$, we can replace the Laplacian regularization with the magnetic Laplacian regularization. Then Equation~\ref{eqn:gsd} can be rewritten as
\begin{equation}\label{eqn:gsdml}
\centering
\begin{split}
\operatorname*{min}_{\mathbf{x}} \{\mu\left\lVert{\mathbf{\bar{x}}-\mathbf{x}}\right\rVert^{2}_{2} + \mathbf{x}^{*}\mathcal{L}_{q}\mathbf{x}\}.
\end{split}
\end{equation}
From Equation~\ref{eqn:gsdml}, We can derive two equivalent closed-form solutions: (1)~Personalized PageRank filter~\cite{10.1145/775152.775191}: $\mathbf{\bar{x}} = \beta{(\mathbf{I}_{n}-\alpha\mathcal{A}_{q})^{-1}}\mathbf{x}$, where $\mathcal{A}_{q}$ is the normalized magnetic adjacency matrix, $\alpha = \frac{1}{\mu+1} \in (0,1)$ and $\beta = \frac{\mu}{\mu+1} \in (0,1)$. (2)~von Neumann kernel~\cite{NIPS2002_778609db}: $\mathbf{\bar{x}} = (\mathbf{I}_{n} + \frac{1}{\mu}\mathcal{L}_{q})^{-1}\mathbf{x}$. The results indicate that an AR$_{K}$ filter or the feedback term of an ARMA$_{F,B}$ filter as $(\mathbf{I}_{n}-\alpha\mathcal{A}_{q})^{-K}$ or $(\mathbf{I}_{n} + \frac{1}{\mu}\mathcal{L}_{q})^{-K}$ has a denoising ability.

\begin{table*}[!t]
\centering
%\vspace{3mm}
\caption{Statistics of datasets}
%\vspace{-2mm}
\begin{tabular}{lcccccccc}
\toprule
\quad &CoRAR &CiteSeerR &PubMed &Cornell &Texas &Washington &Wisconsin\\
\midrule
\#Nodes &2680 &3191 &19717 &195 &187 &230 &265\\
\#Edges &5148 &4172 &44101 &250 &199 &342 &382\\
\#Features &302 &768 &500 &1703 &1703 &1703 &1703\\
\#Classes &7 &6 &3 &5 &5 &5 &5\\
\bottomrule
\end{tabular}
\label{table:stat}
\end{table*}

\begin{table*}[!t]
\centering
\caption{Hyper-parameters of MGC}
%\vspace{-1mm}
\begin{tabular}{lcccccccc}
\toprule
Dataset     &$K$  &$q$  &Learning rate  &$L_{2}$ regularization rate &Dropout rate   &Hidden dimension\\
\midrule
CoRAR       &65     &0      &0.1    &0.001   &0.5     &64\\
CiteSeerR   &30     &0      &0.1    &0.001   &0.5     &64\\
PubMed      &8      &0      &0.1    &0.001   &0.3     &64\\
Cornell     &8      &$\frac{1}{5}$      &0.01   &0.0001     &0.2    &64\\
Texas       &8      &$\frac{1}{4}$      &0.01   &0.0001     &0.4    &64\\
Washington  &8      &$\frac{1}{5}$      &0.01   &0.0001     &0.4    &64\\
Wisconsin   &16     &$\frac{1}{3}$      &0.01   &0.0001     &0.1    &64\\
\bottomrule
\end{tabular}
\label{table:hyper}
\vspace{-2mm}
\end{table*}

\subsection{Graph Neural Network Architecture}
\subsubsection{Graph Augmented Linear Layer}
Based on the graph Fourier transform, a graph convolution layer is defined as $\mathbf{Z} = \sigma{\left(\mathbf{HXW}\right)}$, where $\mathbf{H} \in \mathbb{C}^{n \times n}$ is a graph filter matrix, $\mathbf{X} \in \mathbb{C}^{n \times c_{\mathrm{in}}}$ is a feature matrix, $\mathbf{W} \in \mathbb{C}^{c_{\mathrm{in}} \times c_{\mathrm{out}}}$ is a learnable weight matrix, and $\sigma{(\cdot)}$ is an activation function. 

There are two approaches to simplify a graph convolution layer into a linear layer form, which we call the graph augmented linear layer. The first one, which we name pre-computation style, is defined as $\mathbf{Z} = \sigma{\left(\bar{\mathbf{X}}\mathbf{W}\right)}$, where $\bar{\mathbf{X}} = \mathbf{HX}$ is a $n \times c_{\mathrm{in}}$ matrix multiplied by graph filter with feature matrix before the model training. The other one, which we call pre-prediction style, is defined as $\mathbf{Z} = \sigma{\left(\mathbf{H}\bar{\mathbf{W}}\right)}$, where $\bar{\mathbf{W}}_{i,:} = f_{\theta}(\mathbf{X}_{i,:})$ is a $n \times c_{\mathrm{out}}$ matrix multiplied by feature matrix with a learnable weight matrix. The pre-computation style graph convolution is first proposed in SGC~\cite{pmlr-v97-wu19e}, and the pre-prediction style graph convolution is first proposed in PPNP and APPNP~\cite{klicpera2018combining}. 

Although those two simplified approaches of a graph convolution layer are equivalent, the latter requires backward propagation, which leads to high computational and storage cost in backward propagation stage. In contrast, the computational and storage cost of the pre-computation style graph convolution is eliminated because the graph convolution for this approach does not require backward propagation. Due to the above reasons, we decide to adopt the pre-computation style graph convolution in our models.

We propose an iteration approach to calculate $\mathbf{\bar{X}}=\mathbf{H}_{\mathrm{LR}}\mathbf{X}$ as
\begin{equation}
\centering
\begin{split}
\mathbf{\bar{X}}^{(0)} &= \mathbf{T}^{(0)} = \frac{2}{K+1}\mathbf{X}, \\
\mathbf{T}^{(k+1)} &= \frac{K-k-1}{K-k}\mathbf{P}\mathbf{T}^{(k)}, \\
\mathbf{\bar{X}}^{(k+1)} &= \mathbf{\bar{X}}^{(k)} + \mathbf{T}^{(k+1)}.
\end{split}
\end{equation}
When $k = K-1$, this iteration is terminated.

\subsubsection{Magnetic Graph Convolutional Network}
Based on the architecture of a graph augmented linear layer, we propose a complex-valued GNN named Magnetic Graph Convolutional network (MGC) as 
\begin{equation}
\centering
\begin{split}
\hat{\mathbf{Y}}_{\mathrm{MGC}} = \mathrm{Softmax}{\left(\varsigma\left(\bar{\mathbf{X}}\mathbf{W}^{(0)}\right)\mathbf{W}^{(1)}\right)},
\end{split}
\end{equation}
where $\varsigma(\cdot)$ is a novel complex-valued activation function named Complex Gated Tanh Unit ($\mathbb{C}\mathrm{GTU}$). It is defined as
\begin{equation}
\centering
\begin{split}
\mathbb{C}\mathrm{GTU}(z) &= \mathrm{Tanh}{\left(\mathfrak{R}({z})\right)} \odot \mathrm{Tanh}{\left(\mathfrak{I}({z})\right)},
\end{split}
\end{equation}
where $\mathfrak{R}$ is the real part of a complex-valued tensor and $\mathfrak{I}$ is the imaginary part. When the input is a complex-valued tensor, $\mathbb{C}\mathrm{GTU}$ will turn it into a real-valued tensor. When the input is a real-valued tensor, $\mathbb{C}\mathrm{GTU}$ will be degenerated to Tanh. As \cite{NEURIPS2019_ccdf3864} mentioned, Tanh has a great impact to relieve over-smoothing.

We follow \cite{trabelsi2018deep} for complex-valued weight initialization. Notice that all these settings ensure MGC be degenerated to an RVNN when $q = 0$ or $q = \frac{1}{2}$.
\section{Experiments}
%Baselines
%0. 2-layer MLP
%1. ChebyNet
%2. GCN
%3. SGC
%4. gfNN
%5. PPNP
%6. APPNP
%7. GDC(PPR) with GCN
%8. GDC(HKPR) with GCN
%9. GPR-GNN
%10. S$^{2}$GC
%11. GAT
%12. DiGCN
%13. MagNet

\subsection{Datasets and Experimental Setup}
\noindent\textbf{Datasets.}
We use two different graph datasets, three citation networks, and four webpage networks, for node classification tasks. The detailed statistics of those datasets are shown in Table~\ref{table:stat}. Notice that citation networks are homogeneous graphs and webpage networks are heterogeneous graphs.

\begin{table*}[t]
\centering
\vspace{-1mm}
\caption{Node classification accuracy (\%). OOM means out of GPU memory.}
%\vspace{-1mm}
\resizebox{2.01\columnwidth}{!}{
\begin{tabular}{lccccccc}
\toprule
Model &CoRAR &CiteSeerR &PubMed &Cornell &Texas &Washington &Wisconsin\\
\midrule
ChebyNet    &69.51 ± 1.03 &63.40 ± 1.15 &82.93 ± 0.08 &70.51 ± 3.85 &81.58 ± 2.63 &73.91 ± 2.75 &77.36 ± 2.83\\
GCN         &73.08 ± 0.92 &68.04 ± 0.83 &84.44 ± 0.13 &42.31 ± 1.28 &47.37 ± 1.13 &56.52 ± 2.30 &49.06 ± 1.89\\
SGC         &73.83 ± 0.09 &67.75 ± 0.13 &78.92 ± 0.03 &41.03 ± 0.00 &55.26 ± 0.00 &59.78 ± 1.09 &50.00 ± 0.00\\
PPNP        &78.74 ± 0.57 &67.91 ± 0.59 &84.79 ± 1.10 &70.51 ± 1.28 &80.26 ± 3.95 &80.43 ± 2.17 &76.41 ± 0.91\\
APPNP       &77.77 ± 0.83 &67.87 ± 0.55 &85.15 ± 0.02 &66.67 ± 2.56 &77.63 ± 1.32 &79.35 ± 3.26 &76.42 ± 1.08\\
GDC(PPR)-GCN &78.87 ± 0.35 &68.66 ± 0.96 &82.83 ± 0.18 &65.38 ± 1.22 &73.68 ± 1.95 &80.43 ± 2.07 &74.53 ± 0.90\\
GDC(HKPR)-GCN &77.97 ± 0.29 &\textbf{69.67 ± 0.91} &OOM &46.15 ± 1.73 &55.26 ± 1.04 &56.52 ± 0.31 &50.94 ± 0.12\\
S$^{2}$GC   &\textbf{80.29 ± 0.07} &68.56 ± 0.35 &80.36 ± 0.75 &41.03 ± 0.00 &57.89 ± 0.00 &54.35 ± 0.00 &58.49 ± 0.00\\
DiGCN       &74.07 ± 1.03 &68.06 ± 0.70 &OOM &55.84 ± 0.57 &63.16 + 0.00 &52.17 ± 0.46 &55.68 ± 0.83\\
\hline
\noalign{\vskip 0.5ex}
MagNet      &73.52 ± 0.18 &66.34 ± 1.35 &84.37 ± 0.02 &71.79 ± 1.97 &72.37 ± 1.32 &68.48 ± 1.47 &55.66 ± 2.37\\
\cline{2-8}
Best $q$ &0 &0 &0 &0.2 &0.25 &0.2 &0.15\\
\hline
\noalign{\vskip 0.5ex}
MGC         &79.83 ± 0.44 &68.47 ± 0.30 &\textbf{85.51 ± 0.05} &\textbf{79.49 ± 2.56} &\textbf{84.21 ± 2.63} &\textbf{83.70 ± 1.09} &\textbf{83.02 ± 1.89}\\
\cline{2-8}
Best $q$ &0 &0 &0 &$\frac{1}{5}$ &$\frac{1}{4}$ &$\frac{1}{5}$ &$\frac{1}{3}$\\
\bottomrule
\end{tabular}
}
\label{table:res}
%\vspace{-3mm}
\end{table*}

For citation networks, CoRA, CiteSeer, and PubMed~\cite{Sen_Namata_Bilgic_Getoor_Galligher_Eliassi-Rad_2008} are standard semi-supervised benchmark datasets. In this paper, we adopt CoRAR and CiteSeerR~\cite{zou2019dimensional} instead of CoRA and CiteSeer because the data in both CoRA and CiteSeer are not clean. In CoRA, there exist 32 duplicated papers. In CiteSeer, there exist 161 duplicated papers. Both CoRA and CiteSeer have information leak issue, which means that features include label contexts of papers. More details can be found in Appendix~A of \cite{zou2019dimensional}. In these three citation networks, every node represents a paper and every edge represents a citation from one paper to another. The edge direction is defined from a citing paper to a cited paper. The feature corresponds to the bag-of-words representation of the document and belongs to one of the academic topics. For citation networks, in order to evaluate semi-supervised graph representation learning, we randomly split nodes of each class into 5\%, 10\%, and 85\% for training, validation, and test sets, respectively.

For webpage networks, Cornell, Texas, Washington, and Wisconsin~\cite{lu2003link} are included. Each node represents a webpage and each edge represents a hyperlink between two webpages. The edge direction is from the original webpage to the referenced webpage. The feature of each node is the bag-of-words representation of the corresponding page. For webpage networks, in order to evaluate supervised graph representation learning, we follow previous work~\cite{Pei2020Geom-GCN} to randomly split nodes of each class into 60\%, 20\%, and 20\% for training, validation, and test sets, respectively.

\noindent\textbf{Baselines , detailed setup and hyperparameters.}
To verify the superiority of our model, we introduce ChebyNet~\cite{DBLP:conf/nips/DefferrardBV16}, GCN~\cite{DBLP:conf/iclr/KipfW17}, SGC~\cite{pmlr-v97-wu19e}, PPNP, APPNP~\cite{klicpera2018combining}, GDC(PPR)-GCN, GDC(HKPR)-GCN~~\cite{NEURIPS2019_23c89427}, S$^{2}$GC~\cite{zhu2021simple}, DiGCN~\cite{NEURIPS2020_cffb6e22}, and MagNet~\cite{zhang2021magnet} as baselines. For all these baselines, we use the default setting and parameters as described in the corresponding paper.

We train our models using an Adam optimizer with a maximum of 10,000 epoch and early stopping with patience to be 50. Table~\ref{table:hyper} summaries other hyperparameters of MGC on all datasets. All experiments are tested on a Linux server equipped with an Intel i7-6700K 4.00 GHz CPU, 64 GB RAM, and an NVIDIA GeForce GTX 1080 Ti GPU.

\begin{figure*}[!t]
	\centering
	\subfigure[Cornell]
	{\includegraphics[trim=0 0 40 0,clip,width=0.32\linewidth]{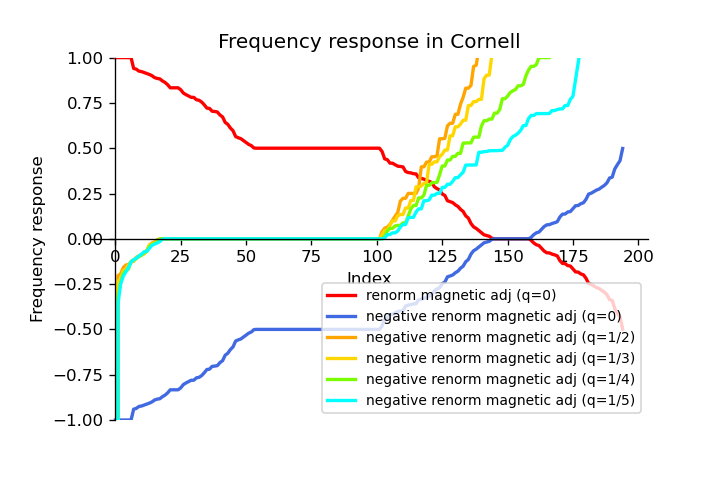}
	}
	\subfigure[Texas]
	{
	\includegraphics[trim=0 0 40 0,clip,width=0.32\linewidth]{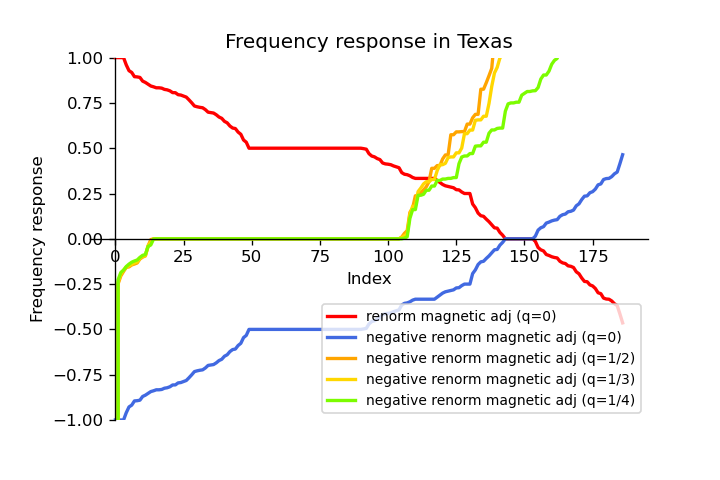}
	}
	\subfigure[Washington]
	{
	\includegraphics[trim=0 0 30 0,clip,width=0.32\linewidth]{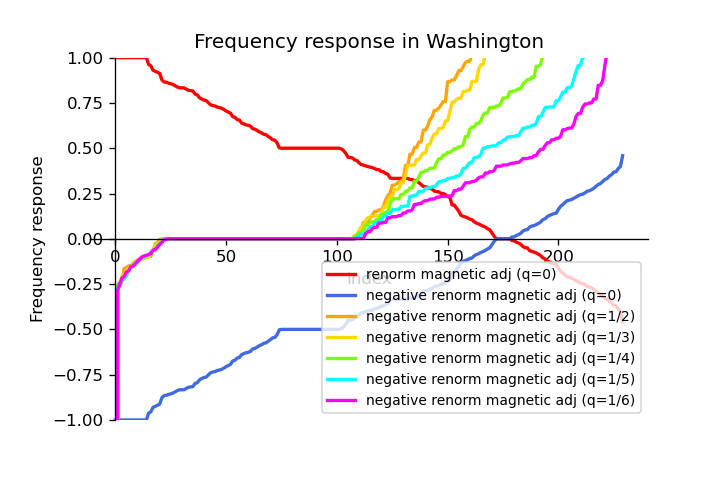}
	}
	\vspace{-3mm}
	\caption{Frequency response in webpage networks w.r.t. $q$}
	\label{fig:FR}
	\vspace{-4mm}
\end{figure*}
\vspace{-1mm}

\subsection{Experiment results and analysis}
Table~\ref{table:res} reports the mean of the node classification accuracy with standard deviation on the test set of each model. In citation networks, MGC is competitive. In PubMed, MGC has the best result among all models. In webpage networks, since the directed graph is cyclic, we have $q \neq 0$. The results in webpage networks show that MGC has a remarkable performance in directed heterogeneous graphs than DiGCN and MagNet.

\subsection{Ablation study}
In order to study which truncated filter is more over-smoothing resistant, we design an ablation experiment. We adopt the S$^{2}$GC as the backbone. The results are in Table~\ref{table:comp}. The experimental results indicate that both Markov Diffusion kernel and LinearRank filter have insignificant performance difference, and the latter is more over-smoothing resistant.

\begin{table*}[t]
\centering
\vspace{-2mm}
\caption{Node classification accuracy (\%) w.r.t. $K$ order}
%\vspace{-1mm}
\begin{tabular}{lccccccccc}
\toprule
Dataset &Filter  &$K = 2$  &$K = 4$  &$K = 8$  &$K = 16$  &$K = 32$  &$K = 64$  &$K = 128$  &$K = 256$\\
\midrule
\multirow{2}{*}{CoRAR}
&MD     &70.85 &74.84 &77.42 &79.26 &\textbf{80.27} &79.26 &74.75 &65.15\\
&LR     &57.91 &64.31 &71.15 &76.68 &78.87 &80.18 &78.56 &74.18\\
\hline
\multirow{2}{*}{CiteSeerR}
&MD     &66.88 &67.73 &\textbf{69.17} &68.91 &69.06 &67.36 &67.84 &66.77\\
&LR     &61.22 &64.95 &67.80 &68.43 &68.72 &68.13 &67.28 &67.36\\
\hline
\multirow{2}{*}{PubMed}
&MD     &79.39 &78.32 &76.67 &74.61 &72.72 &71.97 &71.29 &65.40\\
&LR     &\textbf{80.94} &80.66 &79.70 &77.63 &75.00 &72.85 &72.04 &71.20\\
\bottomrule
\end{tabular}
\label{table:comp}
\vspace{-2mm}
\end{table*}

Note that datasets in webpage networks can be represented as directed cyclic graphs, thus $q \neq 0$. In order to study how the electric changer parameter affects the performance of MGC, we compare $q = 0$ and $q \neq 0$. Recall when $q = 0$, MGC will be degenerated from a CVNN to a RVNN; thus the ablation experiment results also show the comparison between a CVNN and a RVNN. As shown in Table~\ref{table:abla}, when $q \neq 0$, which means the model is a CVNN, MGC has better experiment results. It verifies the effectiveness of the magnetic Laplacian on directed cyclic graphs. Figure~\ref{fig:FR} demonstrates how different values of $q$ affect the frequency response of a GSO. The frequency response of the renormalized magnetic adjacency matrix is described in Appendix~\ref{appendix:a}.

\begin{table}[t]
\centering
%\vspace{3mm}
\caption{Node classification accuracy (\%) w.r.t. $q$}
%\vspace{-1mm}
\begin{tabular}{l|cc}
\toprule
Dataset
&$q = 0$  &$q \neq{0}$ \\
\midrule
Cornell     &76.92 &\textbf{79.49}\\
Texas       &76.32 &\textbf{86.84}\\
Washington  &78.26 &\textbf{82.61}\\
Wisconsin   &75.47 &\textbf{83.02}\\
\bottomrule
\end{tabular}
\label{table:abla}
\vspace{-5mm}
\end{table}

\section{Discussion}
There exist two definitions of the parallel transpose matrix, differing by a minus sign. In \cite{PhysRevE.95.022302}, $\mathbf{T}_{q}(u, v) = \exp\left(i2{\pi}q\left(\mathbf{A}(v, u) - \mathbf{A}(u, v)\right)\right)$. In \cite{FANUEL2018189}, $ \mathbf{T}_{q}(u, v) = \exp\left(i2{\pi}q\left(\mathbf{A}(u, v) - \mathbf{A}(v, u)\right)\right)$. In this paper, we choose the definition of the latter. The normalized magnetic Laplacian which adopts the former parallel transpose matrix is the conjugate transpose of the normalized magnetic Laplacian which adopts the latter. Thus, both magnetic Laplacians share the same eigenvalues. Therefore, if a GNN utilizes the magnetic Laplacian, adopting either definition has no influence on the performance of the GNN.

We notice that some of state-of-the-art GNNs are mini-batch based models. This approach speeds up the training process and allows GNNs to be applied for super-secrecy data. However, determining the value of the electric charger parameter $q$ in each batch is a challenging issue because the value of $q$ depends on the existence of the directed $m$-cycle. If the mini-batch method is based on graph partition, how to partition graphs will directly influence the directed $m$-cycle. If a graph is partitioned inappropriately, each sub-graph may become a directed acyclic graph. In this situation, the value of $q$ for each batch is $0$, which will cause the model to lose the advantage of the magnetic Laplacian. In addition, if the majority of sub-graphs are directed cyclic graphs, and others are directed acyclic graphs, how to decide the value of $q$ for each batch is also an issue.
\vspace{-3mm}
\section{Conclusion and Future Work}
There are plenty of deformed Laplacians which have been researched in graph theory and graph signal processing, such as Chung's directed Laplacian~\cite{chung2005laplacians}, the $p$-Laplacian~\cite{lindqvist2019notes}, the signed Laplacian~\cite{doi:10.1137/1.9781611972801.49}, and the magnetic Laplacian~\cite{PhysRevE.95.022302,FANUEL2018189}. We utilize the property of the magnetic Laplacian for directed graphs and combine it with LinearRank~\cite{10.1145/1148170.1148225} to propose MGC. We test our model in citation networks~\cite{Sen_Namata_Bilgic_Getoor_Galligher_Eliassi-Rad_2008, zou2019dimensional} and webpage networks~\cite{lu2003link}. The experimental results demonstrate that LinearRank is widely applicable for realistic datasets and has a great over-smoothing resistance. MGC is not only well-suited for homogeneous graphs like existing works such as GCN~\cite{DBLP:conf/iclr/KipfW17}, SGC~\cite{pmlr-v97-wu19e}, PPNP, APPNP~\cite{klicpera2018combining}, GDC~\cite{NEURIPS2019_23c89427}, and S$^{2}$GC~\cite{zhu2021simple}, but also effective for heterogeneous graphs. Out of curiosity, we will tap potentials of other spectral ranking algorithms and deformed Laplacians for exploiting the future of graph representation learning.

\section*{Acknowledgements}
Jie Zhang would like to thank Bruno Messias F. Resende who offers the code of the visualization of magnetic eigenmaps, and Hao-Sheng Chen who helps drawing the magnetic eigenmaps.

\bibliography{reference}
\bibliographystyle{icml2022}

%%%%%%%%%%%%%%%%%%%%%%%%%%%%%%%%%%%%%%%%%%%%%%%%%%%%%%%%%%%%%%%%%%%%%%%%%%%%%%%
%%%%%%%%%%%%%%%%%%%%%%%%%%%%%%%%%%%%%%%%%%%%%%%%%%%%%%%%%%%%%%%%%%%%%%%%%%%%%%%
% APPENDIX
%%%%%%%%%%%%%%%%%%%%%%%%%%%%%%%%%%%%%%%%%%%%%%%%%%%%%%%%%%%%%%%%%%%%%%%%%%%%%%%
%%%%%%%%%%%%%%%%%%%%%%%%%%%%%%%%%%%%%%%%%%%%%%%%%%%%%%%%%%%%%%%%%%%%%%%%%%%%%%%
\newpage
\appendix
\onecolumn
\section{The frequency response of the normalized magnetic adjacency matrix and the renormalized magnetic adjacency matrix}
\label{appendix:a}

Since the symmetric normalized Laplacian and the random walk normalized Laplacian share same eigenvalues~\cite{von2007tutorial}, we can derive that symmetric normalized magnetic Laplacian and random walk normalized magnetic Laplacian share same eigenvalues. We denote the set of eigenvalues of normalized Laplacian in ascending order as $\lambda$ and the set of eigenvalues of normalized magnetic Laplacian in ascending order as $\lambda_{q}$. Both $\lambda$ and $\lambda_{q}$ are named frequency response in graph signal processing. Therefore, the frequency response of the normalized magnetic adjacency matrix can be written as $1 - \lambda_{q}$. In~\cite{PhysRevE.95.022302}, it has been proved that $0 = \lambda^{(0)} \le \lambda^{(0)}_{q}$, where $\lambda^{(0)}$ and $\lambda^{(0)}_{q}$ represent the lowest eigenvalues of the normalized Laplacian and the normalized magnetic Laplacian, respectively. Note that when $q = 0$, the equality is achieved. We assume that 
\begin{equation}\label{eqn:deg}
\centering
\begin{split}
\mathbf{D}_{s} \approx \bar{d}\mathbf{I}_{n}, 
\end{split}
\end{equation}
where $\bar{d}$ represents average node degree. It can yield that
\begin{equation}\label{eqn:adj}
\centering
\begin{split}
\mathbf{A}_{s} \approx \bar{d}\left(\mathbf{I}_{n} - \mathcal{L}\right).
\end{split}
\end{equation}
Then, we put Equations~\ref{eqn:deg} and~\ref{eqn:adj} into the renormalized adjacency matrix as
\begin{equation}
\centering
\begin{split}
\widetilde{\mathcal{A}} &= \widetilde{\mathbf{D}}^{-\frac{1}{2}}_{s}\widetilde{\mathbf{A}}_{s}\widetilde{\mathbf{D}}^{-\frac{1}{2}}_{s} \\
&= \left(\mathbf{D}_{s}+\mathbf{I}_{n}\right)^{-\frac{1}{2}}\left(\mathbf{A}_{s}+\mathbf{I}_{n}\right)\left(\mathbf{D}_{s}+\mathbf{I}_{n}\right)^{-\frac{1}{2}} \\
&\approx \left(\bar{d}\mathbf{I}_{n}+\mathbf{I}_{n}\right)^{-\frac{1}{2}}\left(\bar{d}\left(\mathbf{I}_{n}-\mathcal{L}\right)+\mathbf{I}_{n}\right)\left(\bar{d}\mathbf{I}_{n}+\mathbf{I}_{n}\right)^{-\frac{1}{2}} \\
&= \mathbf{I}_{n} - \frac{\bar{d}}{\bar{d}+1}\mathcal{L}
\end{split}
\end{equation}
Adopting the same method, we can derive $\widetilde{\mathcal{A}}_{q} \approx \mathbf{I}_{n} - \frac{\bar{d}}{\bar{d}+1}\mathcal{L}_{q}$. Hence, the frequency response of the renormalized adjacency matrix is approximated as $1 - \frac{\bar{d}}{\bar{d}+1}\lambda$, the frequency response of the renormalized magnetic adjacency matrix is approximated as $1 - \frac{\bar{d}}{\bar{d}+1}\lambda_{q}$. Likewise, the frequency response of the negative renormalized magnetic adjacency matrix is approximated as $\frac{\bar{d}}{\bar{d}+1}\lambda_{q} - 1$.
%%%%%%%%%%%%%%%%%%%%%%%%%%%%%%%%%%%%%%%%%%%%%%%%%%%%%%%%%%%%%%%%%%%%%%%%%%%%%%%
%%%%%%%%%%%%%%%%%%%%%%%%%%%%%%%%%%%%%%%%%%%%%%%%%%%%%%%%%%%%%%%%%%%%%%%%%%%%%%%

\end{document}